\documentclass[10pt,letterpaper,journal,twoside]{IEEEtran}
\usepackage{paper-support}

\title{Ferrofluid Shape Control\\
with Deep Reinforcement Learning}
\author{Anthony~Czubarow, Antonio~Terpin, and Raffaello~D'Andrea%
\thanks{All authors are with the Institute for Dynamic Systems and Control, ETH
Zürich, Zürich, Switzerland.}}
\IEEEaftertitletext{%
  \noindent\begin{minipage}{\textwidth}
    \centering
    \includegraphics[width=\textwidth]{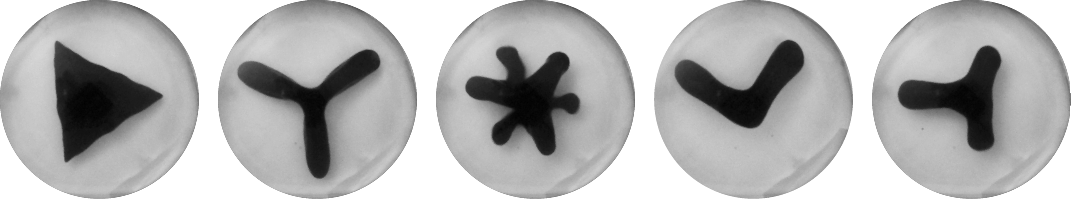}
    \captionof{figure}{Ferrofluid shapes reached through deep reinforcement learning.}
    \label{fig:title-teaser}
  \end{minipage}\par
  \vspace{0.8\baselineskip}
}

\begin{document}
\maketitle
\begin{abstract}
Ferrofluids are liquids containing magnetic nanoparticles, enabling shape control
through external magnetic fields. {This deformability supports
their use as soft robotic bodies that can navigate confined spaces and
manipulate objects.}
Their nonlinear continuum dynamics, however, make reliable control challenging.
Existing work shows how {spatial magnetic fields
from electromagnet arrays} can establish and maintain desired ferrofluid
configurations. A complementary objective is to exploit the
fluid's dynamics under limited actuation to reach prescribed shapes
that may not be sustainable at equilibrium.
Here, we use deep model-based reinforcement learning to control these transient
shapes. Learning directly through physical interaction, our policies select
current commands for a single fixed electromagnet based on visual observations,
guiding the moving ferrofluid toward prescribed geometries.
The learned policies repeatedly attain a variety of shapes spanning polygons,
asterisks and pentominoes. These results demonstrate that learning directly from
physical interaction enables reliable fluid shaping,
opening new possibilities for controlling ferrofluid-based soft robots.
\end{abstract}


\section{Introduction}
\label{sec:introduction}
Ferrofluids are liquids that contain magnetic nanoparticles and deform
under applied magnetic fields \citep{socoliuc2022}. Their ability to
stretch, branch and reorganize supports their use in soft
robotics. Ferrofluid robots form structures that capture and transport objects
\citep{fan2020}, split and recombine to negotiate confined passages
\citep{fan2022scale}, and exploit wetting to create mobile droplets and
reconfigurable liquid structures \citep{sun2022}.
{Controlled changes in ferrofluid geometry can
generate locomotion through cycles of elongation and contraction
\citep{kong2021} or through the nonreciprocal strokes of liquid cilia
\citep{sun2022}.}
{These capabilities motivate control of the liquid's
geometry for soft robotic interactions with the surroundings.}
{Additional examples are reviewed in \cref{sec:related-work}.}

Existing shape-control methods connect magnetic commands to geometric
descriptors or desired contours. Visual feedback regulates elongation and
orientation \citep{ahmed2021}, as well as deformation ratio
\citep{kong2021}; Bayesian optimization also tunes feedback controllers for
three-dimensional position, stretch direction and stretch radius
\citep{ahmed2024}. For full silhouettes, learned models of water--ferrofluid
interfaces guide the selection of coil currents by predicting the target match
at steady state \citep{alessandri2022}. Related work shapes
nonmagnetic droplets supported by an air--ferrofluid interface: electromagnet
arrays produce convex, concave and time-varying contours
\citep{harischandra2023shaping}, learned models predict shape sequences and
infer successive actuations \citep{harischandra2023learning}, and physical
Bayesian optimization selects coil settings for requested outlines
\citep{harischandra2025}. These studies establish several routes to geometric
control, including configurations maintained under selected magnetic fields.

{Fluid motion offers a further source of attainable geometry,
much as learned dynamic maneuvers extend the reach of soft robot arms
\citep{jitosho2023agile}.} A moving
interface can pass through shapes that it cannot sustain at equilibrium.
In immersed drops, interfacial tension supplies a restoring stress, inertia
sustains motion and viscosity damps the response
\citep{miller1968oscillations,trinh1982small}.
{Magnetic actuation influences both how the fluid deforms
and which geometries emerge. Deformation followed by field removal produces
oscillation and relaxation \citep{backholm2017}, while magnetic preparation
selects spreading patterns \citep{chen2010} and a bottom electromagnet produces
noncircular thin-drop geometries \citep{li2024patterns}.}
{Prescribed magnetic pulses also drive confined ferrofluid drops
through transient nonelliptical outlines to pump surrounding liquid
\citep{wu2023actuator}.} Mechanically excited
ferrofluid puddles exhibit star-shaped modes whose frequencies can be tuned
magnetically \citep{jamin2016}. These results motivate using the evolving
fluid response to create chosen silhouettes. The control problem is to select
magnetic commands as the fluid moves so that a prescribed shape appears
along its trajectory, even if only briefly.

{Machine learning offers tools for controlling the
nonlinear, time-dependent behavior of soft robots
\citep{kim2021softreview,laschi2023learning}. Reinforcement learning, in
particular, improves control policies through interaction guided by task
rewards \citep{sutton2018}. It has been used to exploit viscous coiling to deposit prescribed curves,
with nozzle motions learned in simulation and executed physically
\citep{chaudhary2023}, and to control plasma shape in a tokamak with
policies trained in simulation \citep{degrave2022}.}
{Simulation pretraining followed by physical adaptation
enables ultrasound-driven microrobot navigation \citep{medany2025}, while
learning directly from camera observations regulates interface position and
droplet size in microfluidic experiments \citep{dressler2018}.}

{For ferrofluids, the challenge of modeling coupled
magnetic and fluid dynamics \citep{alessandri2022} motivates learning directly
from physical interaction, avoiding the need to construct a fluid simulator
{and the challenges of sim-to-real transfer.}
DreamerV3 supports data-efficient learning by
fitting a world model to collected experience and training a policy through
imagined interaction with that model \citep{hafner2025}.} Learning through
such models has been demonstrated in simulated visual-control tasks
\citep{hafner2020dreamer} and directly on physical robots \citep{wu2023},
on a physical labyrinth game \citep{bi2024}, and in active-flow-control
experiments {\citep{terpin2026,bonomi2026,terpin2026artpiv}}.

\begin{figure}[!t]
\centering
\includegraphics[trim=0 45bp 0 45bp,clip,width=\linewidth]{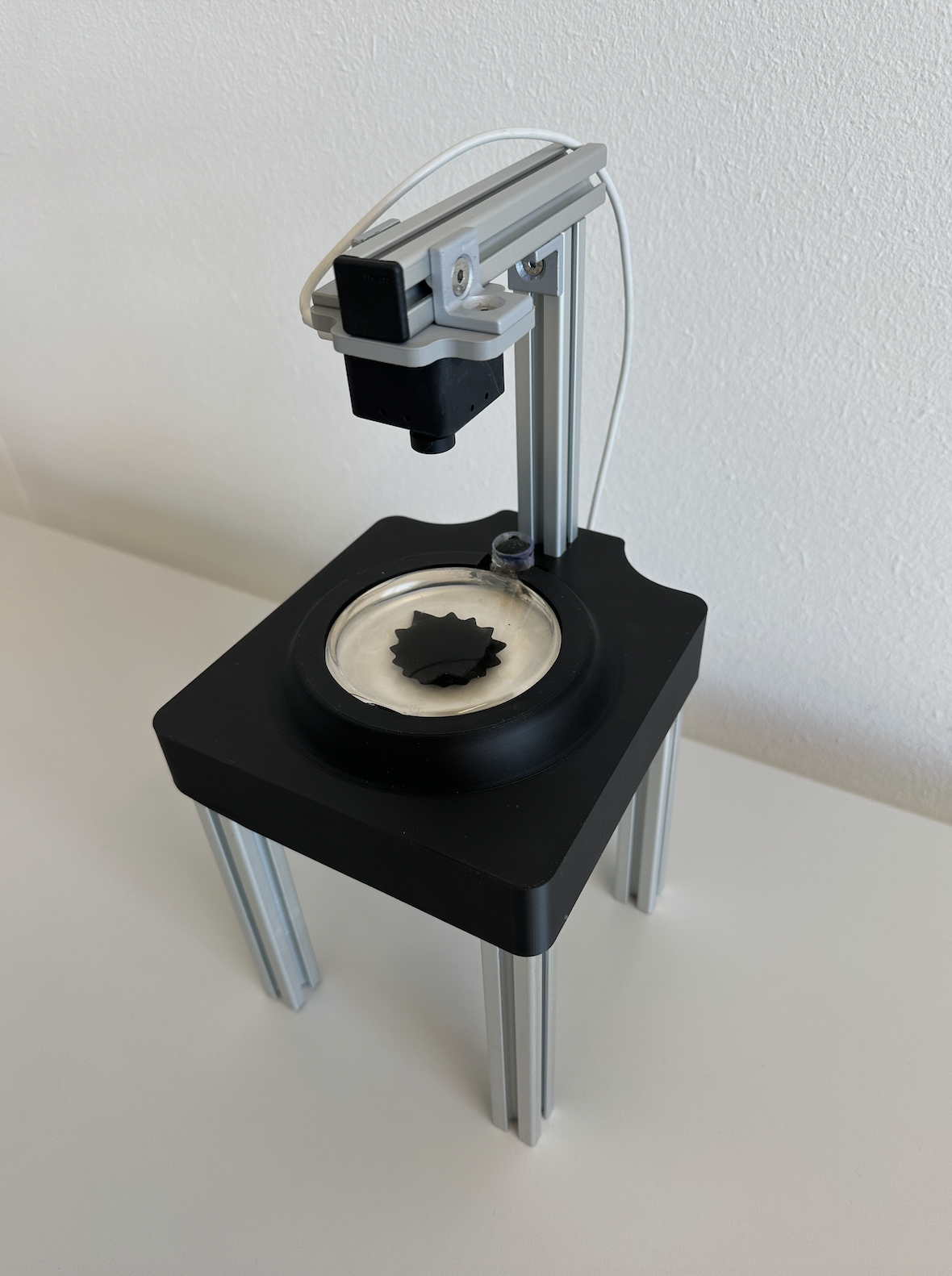}
\caption{Experimental platform for ferrofluid shaping. A downward-facing
camera observes the fluid in a glass jar. A 3D-printed case supports the jar,
and an item-profile frame holds the camera. The shaping electromagnet sits
beneath the jar. \Cref{sec:platform} describes the internal arrangement.}
\label{fig:testbed}
\end{figure}

{Here, we use deep model-based reinforcement learning
to exploit the dynamics of ferrofluid immersed in water for prescribed shape
control (Fig.~\ref{fig:title-teaser}). We learn a dynamics model from physical
experience and use it to train policies, which select current
commands for a single fixed electromagnet (\cref{fig:testbed,fig:setup}).}
{The same approach to learning
directly on hardware could be explored for ferrofluid locomotion and
manipulation.}

\Needspace{2\baselineskip}
\noindent\textbf{Contributions.}
{We introduce a physical platform
designed to be easy to reproduce and operate, combining a single fixed
electromagnet and visual feedback.
{We integrate DreamerV3~\citep{hafner2025} with the platform to learn
shape-control policies directly through physical interaction. We formulate
a reward for improvements in the best window-averaged silhouette overlap
and demonstrate its effectiveness for learning transient shape control.}
We evaluate transient shape control for 20 silhouettes
spanning polygons, asterisks with three to six rays and pentominoes.
We compare trained and untrained policies in shape accuracy,
attainment frequency and time to attainment, and characterize persistence
near each episode's peak intersection-over-union (IoU) score.}

\section{Experimental System}
\label{sec:experimental-system}
\subsection{Physical platform and system overview}
\label{sec:platform}
\begin{figure}
\centering
\includegraphics[width=\linewidth]{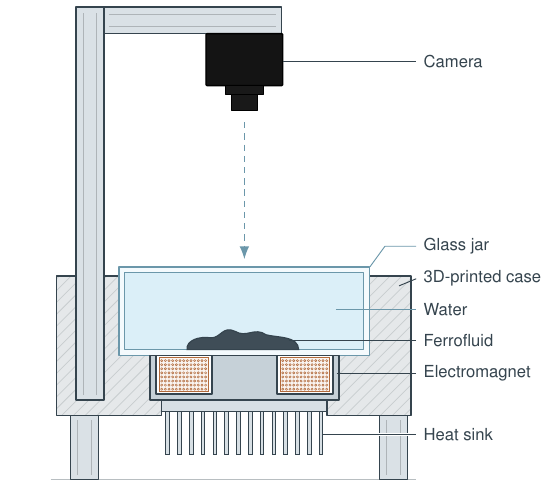}
\caption{Schematic cutaway of the apparatus (not to scale). The jar rests above
the electromagnet; the cutaway shows its central pole and annular winding.
The heat sink sits below the magnet.}
\label{fig:setup}
\end{figure}

The experimental platform couples visual shape feedback to electromagnetic actuation.
It consists of ferrofluid immersed
in water in a glass jar, an overhead camera and a single fixed electromagnet centered beneath
the jar (Fig.~\ref{fig:testbed}). \Cref{fig:setup} shows the internal arrangement.
{We use the glass jar, ferrofluid, surrounding water and electromagnet from
a Dancing Ferrofluid Speaker Mate Music Rhythm
Lamp~\cite{sovenomundDancingMate}. The jar is 85~mm in diameter and 30~mm {tall},
with approximately 2~mm-thick glass. The closed jar contains no air
headspace. The jar contains 5~mL of ferrofluid, which rests on the inner
glass base, approximately 2~mm above the electromagnet face.}
{A 3D-printed case houses the jar and electromagnet. An item-profile
frame supports the downward-facing camera {with its lens
80~mm above the jar's top glass surface}, and a heat sink beneath the magnet
limits heating, keeping the fluid at approximately room temperature during
operation.}

The image-based closed-loop controller runs at 55~Hz.
{The host computer processes camera images and sends current
set points to an embedded controller. The embedded controller regulates coil
current using local current-sensor feedback.}
This arrangement separates image-based shape control from local current
regulation.

\subsection{Visual sensing and shape measurement}
\label{sec:visual-sensing}
We use camera images {as input to the policy} and to measure agreement with
the target silhouette.
An e-con Systems See3CAM\_24CUG camera with a global shutter acquires the
images. We place the platform in an enclosed space to keep
lighting conditions constant during data collection.

We convert the images to grayscale, restrict them to a fixed circular region
of interest and resize them to $64\times64$ pixels for policy input. Intensity
thresholding at this resolution produces binary ferrofluid masks for shape
scoring, isolating the fluid from the white background and removing the internal
contrast of the camera reflection visible in the photographs.
Section~\ref{sec:control-formulation} defines the shape score and
temporal reward, and Section~\ref{sec:learning-implementation} describes registration.
The policy thus observes the fluid image, while shape scoring operates on the
extracted silhouette.

\subsection{Electromagnetic actuation and characterization}
\label{sec:actuation}
The electromagnet exerts magnetic forces on the fluid in response to the
commanded current.
{The shaping electromagnet has a steel body with a
22~mm-diameter central pole and a casing with outer and inner diameters of
64~mm and 60~mm, respectively. Its overall height is 15~mm. Enamelled wire of
0.5~mm outer diameter fills the annular space between the pole and casing
(Fig.~\ref{fig:setup}), with thermal paste between the windings
to improve heat conduction.}

A Teensy~4.1 microcontroller regulates coil current at 1~kHz using
proportional--integral (PI) control. A BTS7960
H-bridge drives the coil, while an M5Stack Unit INA226-1A sensor with a
$0.08\,\Omega$ shunt provides current feedback. We validate the current using
a calibrated oscilloscope.

To estimate the coil inductance, we bypass PI regulation and apply
fixed-duty voltage pulses. The INA226 records the resulting current rise,
which we fit to $i(t)=b+A(1-e^{-t/\tau})$, where $t$ is time since pulse onset, $b$ is the current offset
and $A$ the rise amplitude. The fits yield a median time constant of
$\tau=4.125\,\mathrm{ms}$. Using the coil resistance $R=25\,\Omega$ measured with a multimeter, we estimate the inductance with the first-order coil model:
$L_{\mathrm{coil}}=R\tau\approx103\,\mathrm{mH}$.

\begin{figure}[!t]
\centering
\includegraphics[width=\linewidth]{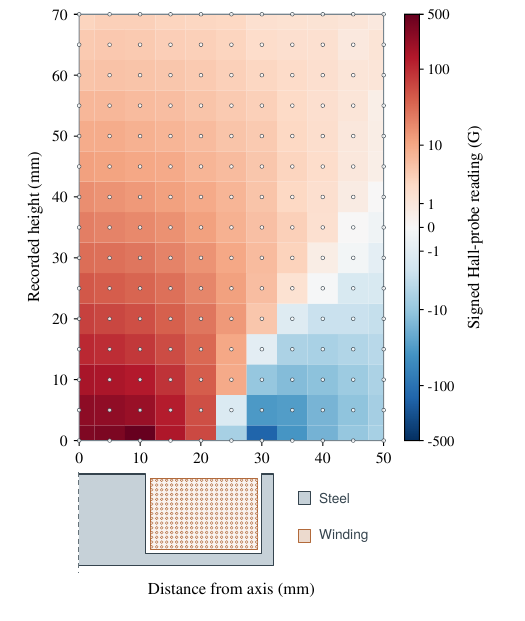}
\caption{Hall-probe readings in a plane through the magnet axis at 0.5~A.
Points mark sampled positions. The signed logarithmic color scale gives
readings in gauss and retains their polarity. The schematic magnet
half-section below shares the map's radial scale.}
\label{fig:magnetic-field}
\end{figure}

\Cref{fig:magnetic-field} summarizes the spatial Hall-probe readings at
0.5~A. We position an {AlphaLab GM1-ST gaussmeter} with a 3D-printed
slotted jig and sample a plane through the magnet axis on a 5~mm grid.
{The inductance and field map characterize the actuator's
electrical response and spatial magnetic field.}
The controller selects current commands from visual observations to produce
a prescribed ferrofluid silhouette while the electromagnet remains fixed in
space.

\begin{figure*}[t]
\centering
\includegraphics[width=\textwidth]{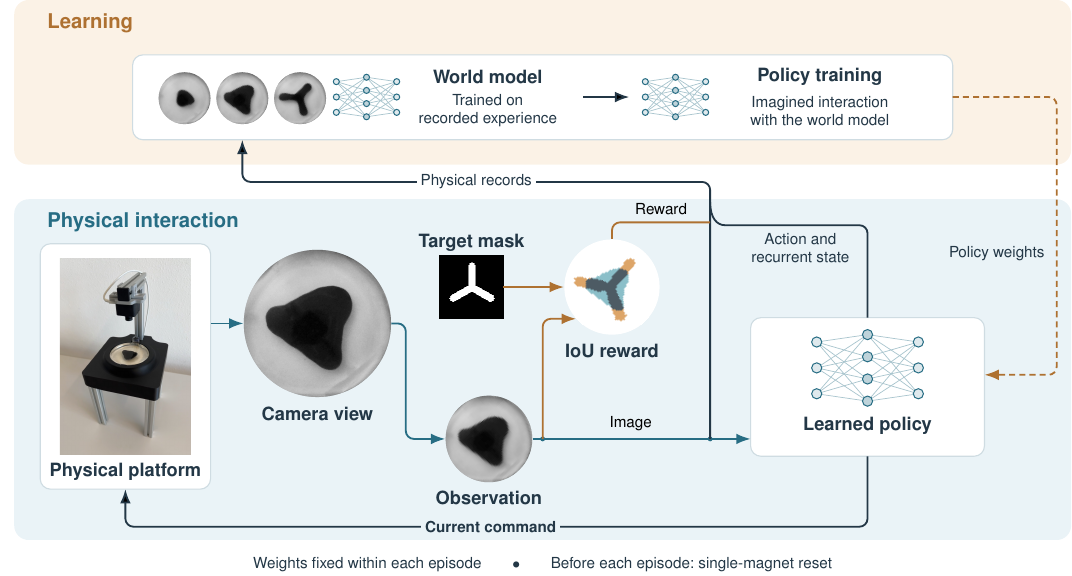}
\caption{Application of DreamerV3~\citep{hafner2025} to ferrofluid shaping.
Recorded images, actions and rewards train the world model.
The policy is trained through imagined interaction with this model.
{The control loop runs at 55~Hz.}
The target mask enters the IoU record improvement reward
path. Learning and interaction run concurrently. The controller receives updated policy weights before each episode (dashed arrow), then holds them
fixed throughout it. The thumbnails show a recorded trajectory
sampled every ten control steps at $64\times64$ resolution.
The overlap image shows the intersection (dark),
target-only (orange) and observation-only (blue) regions.}
\label{fig:learning-loop}
\end{figure*}

\section{Method}
\label{sec:methods}
\subsection{Reinforcement learning}
\label{sec:rl-formulation}
\begin{figure}[!t]
\centering
\includegraphics[width=\linewidth]{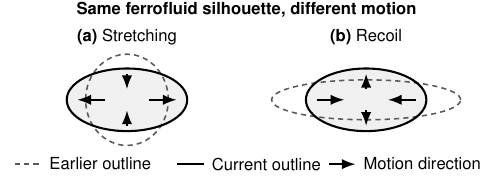}
\caption{{Schematic ferrofluid silhouettes during stretching
(a) and recoil (b). Solid lines show the same current outline, while dashed
lines show earlier outlines and arrows indicate motion, illustrating how
identical silhouettes can evolve differently.}}
\label{fig:transient-dynamics}
\end{figure}

We formulate shaping the ferrofluid toward a prescribed silhouette as an
episodic partially observable Markov decision process \citep{kaelbling1998}.
At step $k$, a policy $\pi$ uses camera observations $y_k$ to select a scalar
action $a_k\in[-1,1]$, which sets the current of the fixed electromagnet
(Section~\ref{sec:learning-implementation}). We assume Markov evolution for
the complete physical state $x_k$, comprising the fluid interface, motion of
both liquids, and the internal actuator and sensing states.
{A single camera image $y_k$ does not determine $x_k$:
the same ferrofluid silhouette can accompany different fluid motion
(\cref{fig:transient-dynamics}).}

Each episode comprises $T$ current commands at a nominal rate of 55~Hz.
We train $\pi$ directly on the
physical platform using DreamerV3 \citep{hafner2025} to maximize the expected
discounted return
$\mathbb{E}_{\pi}[\sum_{k=0}^{T-1}\gamma^k r_{k+1}]$ \citep{sutton2018},
where $\gamma\in[0,1]$ is the discount factor and $r_{k+1}$ rewards an
improvement in the best target match achieved during the episode. We define
this reward and the memory that augments $x_k$ below.

\subsection{Shape reward and augmented state}
\label{sec:control-formulation}
We reward improvements in the best match achieved over a short window of
observations. This rewards a visit to the target even when the fluid
subsequently moves away.

{We measure shape agreement by IoU.} Let $M^\star$
denote the target foreground and $M_k$ the ferrofluid foreground in each
post-action observation $y_k$, $k\geq1$. After aligning the target to the
observation by translation,
rotation and uniform scaling, let $\widehat M^\star_k$ denote the registered
target on the same image canvas. The shape score is
\begin{equation*}
 s_k=|M_k\cap\widehat M^\star_k|\,/\,|M_k\cup\widehat M^\star_k|.
\end{equation*}
Here, $|\cdot|$ counts foreground pixels. The score ranges from zero for no
overlap to one for an exact match. Registration reduces sensitivity to
the drop's position, orientation and size within the search ranges of
Section~\ref{sec:learning-implementation}, so the score compares shape rather than fixed image pose;
that section describes the numerical registration.

To score a short visit to the target, we average over $L$ consecutive
observations. We start the score history with $y_1$, the first observation
following an action, and exclude the initial observation $y_0$. The window
mean $\bar s_k$ and best mean so far $m_k$ are, for $k\geq L$,
\begin{equation*}
 \bar s_k=\frac{1}{L}\sum_{j=k-L+1}^{k}s_j,
 \qquad m_k=\max(m_{k-1},\bar s_k),
\end{equation*}
with $m_k=0$ for $k<L$.
The reward following action $a_k$ is the improvement in this record,
$r_{k+1}=m_{k+1}-m_k$. The first complete window establishes the initial
rewarded record; subsequent rewards occur only when the record improves.

To retain the history needed for this reward, let
$w_k=(s_{k-L+2},\ldots,s_k)$ store the most recent $L-1$ scores, with
$s_j=0$ for $j\leq0$. The environment state is
\begin{equation*}
 z_k=(x_k,w_k,m_k,k).
\end{equation*}
After action $a_k$, the next image supplies $s_{k+1}$. Once $k+1\geq L$,
averaging this score with the entries of $w_k$ gives $\bar s_{k+1}$.
We update $m_{k+1}$ as above and retain the latest $L-1$ scores in $w_{k+1}$.
The clock determines when the first window is complete and when the episode
ends at $T$. These updates make $z_k$ a Markov state for both the physical
evolution and the reward. At reset, $w_0$ is zero, $m_0=0$ and $k=0$.
The target is fixed for each run. The policy observes only camera images;
the reward memory and clock remain internal to the environment.

For $T\geq L$, these increments telescope:
\begin{equation*}
 \sum_{k=0}^{T-1}r_{k+1}=m_T=\max_{L\leq k\leq T}\bar s_k.
\end{equation*}
The undiscounted return thus measures the best window-averaged match.
Discounting additionally favors earlier improvements. Repeatability and
uninterrupted time near the target remain separate evaluation quantities.

\subsection{Physical implementation}
\label{sec:learning-implementation}
\paragraph{Observations and actions}
Each observation $y_k$ is the $64\times64$, 8-bit grayscale image from
Section~\ref{sec:visual-sensing}.
We clip each scalar action to $[-1,1]$ and map the resulting value $a_k$
to the absolute current set point
$I_k^{\mathrm{cmd}}=0.25(a_k+1)\,\mathrm{A}$, spanning $0$--$0.5\,\mathrm{A}$.
The embedded regulator tracks this set point until the next command or reset
(Section~\ref{sec:actuation}).

\paragraph{Episode reset}
Before each episode, we command the same central electromagnet to
$0.8\,\mathrm{A}$ for 8~s to recombine any separated ferrofluid droplets
into a single drop. This reset is separate
from the $T$ learned shaping actions. We also reset the reward history and
the agent's recurrent state.

\paragraph{Shape registration}
We represent targets as binary masks with white foreground on a black
background and resize them to $64\times64$ using nearest-neighbor
interpolation. For each observation, we first align the foreground centroids
by translation. We then approximate the maximum IoU by a coarse search over
uniform scales in $[0.5,2]$ and rotations in $[-180^\circ,180^\circ]$ about
the target centroid. We rasterize each transformed target on the image
canvas. Lower-scoring matches undergo local refinement of translation,
scale and rotation. The final transform defines $\widehat M^\star_k$ and
the score $s_k$, which we evaluate within the image canvas. Frames with
fewer than 5 foreground pixels receive zero.

\paragraph{Learning configuration}
We train a separate policy for each target.
We use DreamerV3 with the \texttt{size25m} preset and a convolutional neural network
to encode the images. We set the discount factor to $\gamma=332/333$ and
the reward window to $L=7$ observations.
Learning and policy inference run concurrently (Fig.~\ref{fig:learning-loop}),
with one graphics processing unit assigned to each task.

\begingroup
\paragraph{Training and evaluation}
For each target, we train three independent policies with three seeds for
$N$ episodes of $T$ steps, giving $NT$ training steps per policy; each step
applies one shaping action, and resets are excluded from this count.
{We evaluate freshly initialized, untrained
stochastic policies for 50 episodes per target and seed, using the same target,
reset and horizon with weights fixed.}
We evaluate the final checkpoint of each run for 50 episodes using
the same target, reset and horizon, with weights fixed and actions sampled
from the policy. The numbers of training episodes and steps for each target
family are given in
\cref{sec:polygons,sec:asterisks,sec:pentominoes}.
\endgroup

\Needspace{8\baselineskip}
\section{Results}
\label{sec:results}
\begingroup
\subsection{Evaluation overview}
\label{sec:physical-learning}
We study 20 target shapes in three families: 4 polygons, 4 asterisks
and 12 pentominoes. {For each family, we report episode return and
attainment-time curves for three seeds, alongside 50 evaluation rollouts of
each untrained and final policy. The figures pair each target with illustrative
ferrofluid images and show how performance develops over training steps.}

{We report undiscounted episode return, the highest seven-observation
mean of the IoU score defined in \cref{sec:control-formulation} and illustrated
in \cref{fig:circular-reference}.} Learning curves show the three training
runs and their mean, smoothed with a trailing 10-episode average.
{\Cref{tab:three-seed-summary} compares untrained and final
policies over 50 evaluation episodes per seed. We average unsmoothed returns
within each seed, then report the mean and sample
{standard deviation (std)} across
the three seed averages. {Early and late training refer to the
first and last 50 training episodes per seed, respectively.}}
{Final-policy mean returns exceed untrained-policy means
for all 20 targets. Mean return increases from 0.581 to 0.760, a relative gain
of 30.7\% when returns are averaged across all shapes.}
{{The large photographs show each target's highest logged
single-frame IoU across training and final-policy evaluation.}} {Smaller
photographs illustrate lower and intermediate IoUs.}

\begin{table}[!t]
\caption{{Untrained- and final-policy episode returns}}
\label{tab:three-seed-summary}
\centering
\small
\setlength{\tabcolsep}{3pt}
\renewcommand{\arraystretch}{0.97}
{\color{black}%
\begin{tabular*}{\linewidth}{@{\extracolsep{\fill}}lcc@{}}
\toprule
Target & {Untrained policy} & {Final policy} \\
\midrule
\multicolumn{3}{l}{\textit{Polygons}} \\
Triangle & {$0.757 \pm 0.001$} & $0.868 \pm 0.020$ \\
Square & {$0.857 \pm (<0.001)$} & $0.925 \pm 0.009$ \\
Pentagon & {$0.904 \pm (<0.001)$} & $0.916 \pm 0.003$ \\
Hexagon & {$0.931 \pm (<0.001)$} & $0.941 \pm (<0.001)$ \\
\midrule
\multicolumn{3}{l}{\textit{Asterisks}} \\
3 rays & {$0.372 \pm (<0.001)$} & $0.658 \pm 0.073$ \\
4 rays & {$0.421 \pm (<0.001)$} & $0.597 \pm 0.028$ \\
5 rays & {$0.469 \pm (<0.001)$} & $0.529 \pm 0.009$ \\
6 rays & {$0.540 \pm (<0.001)$} & $0.613 \pm 0.042$ \\
\midrule
\multicolumn{3}{l}{\textit{Pentominoes}} \\
F & {$0.599 \pm 0.002$} & $0.742 \pm 0.029$ \\
I & {$0.331 \pm 0.001$} & $0.853 \pm 0.009$ \\
L & {$0.411 \pm 0.002$} & $0.745 \pm 0.005$ \\
N & {$0.489 \pm 0.002$} & $0.735 \pm 0.024$ \\
P & {$0.769 \pm 0.002$} & $0.849 \pm 0.018$ \\
T & {$0.546 \pm (<0.001)$} & $0.768 \pm 0.021$ \\
U & {$0.568 \pm 0.001$} & $0.715 \pm 0.042$ \\
V & {$0.489 \pm (<0.001)$} & $0.734 \pm 0.019$ \\
W & {$0.510 \pm 0.002$} & $0.737 \pm 0.008$ \\
X & {$0.647 \pm (<0.001)$} & $0.818 \pm 0.016$ \\
Y & {$0.511 \pm 0.001$} & $0.747 \pm 0.027$ \\
Z & {$0.499 \pm (<0.001)$} & $0.701 \pm 0.029$ \\
\bottomrule
\end{tabular*}
}
\par\smallskip
\parbox{\linewidth}{\footnotesize {Mean $\pm$ sample {std} across
three policy means, each computed over 50 evaluation episodes.
{A std} below 0.001 is shown as $<0.001$.}}
\end{table}

{To assess whether learning makes target shapes appear more
often and earlier within an episode, we use a shape-specific threshold.
Let $D$ be the highest IoU
of a filled circle aligned to that target, as illustrated in
\cref{fig:circular-reference}. We set $q=D+\alpha(1-D)$ with $\alpha=0.5$,
halfway between this circular overlap and a perfect match. An episode attains
the threshold at the first observation with $s_k\geq q$. This criterion uses
single-frame IoU; the training reward retains its seven-observation window.}

{The lower curves in
\cref{fig:polygons-results,fig:asterisk-results,fig:pentomino-results-one,fig:pentomino-results-two}
show capped first-attainment time: the recorded elapsed time of that crossing,
or the episode's recorded end time if no crossing occurs.
{Recorded times exclude reset actuation but include
episode-start overhead before the first scored action.}
A plus sign marks each unsuccessful episode. We apply the same
10-episode smoothing and three-seed averaging as for returns, keeping all
episodes in the calculation. A decrease therefore reflects more frequent
attainment, earlier crossings, or both. The threshold is an operational IoU
criterion; \cref{fig:attainment-examples} in \cref{sec:attainment-examples}
shows example training photographs just above each threshold.}

{\Cref{tab:attainment-summary} reports final-policy
crossings and capped times at the selected thresholds. No crossings occurred
in the 3\,000 untrained-policy evaluation episodes at these thresholds.
\Cref{sec:threshold-sensitivity} examines sensitivity to the threshold choice.}

\begin{table}[!t]
\caption{{Final-policy threshold crossings and timing}}
\label{tab:attainment-summary}
\centering
\footnotesize
\setlength{\tabcolsep}{2.2pt}
\renewcommand{\arraystretch}{0.98}
{\color{black}%
\begin{tabular*}{\linewidth}{@{\extracolsep{\fill}}lcc|lcc@{}}
\toprule
{Target} & {Crossings} & {Time (s)} & {Target} & {Crossings} & {Time (s)} \\
\midrule
Triangle & 122/150 & $2.35 \pm 0.86$ & L & 143/150 & $4.03 \pm 1.09$ \\
Square & 127/150 & $3.05 \pm 0.25$ & N & 134/150 & $4.64 \pm 0.93$ \\
Pentagon & 0/150 & \textemdash & P & 86/150 & $7.13 \pm 2.73$ \\
Hexagon & 0/150 & \textemdash & T & 128/150 & $3.66 \pm 1.61$ \\
3 rays & 98/150 & $3.02 \pm 0.80$ & U & 52/150 & $8.80 \pm 2.50$ \\
4 rays & 28/150 & $4.24 \pm 0.19$ & V & 116/150 & $5.68 \pm 1.31$ \\
5 rays & 0/150 & \textemdash & W & 138/150 & $5.05 \pm 0.30$ \\
6 rays & 5/150 & $4.53 \pm 0.03$ & X & 125/150 & $5.51 \pm 0.77$ \\
F & 80/150 & $7.30 \pm 2.25$ & Y & 128/150 & $5.23 \pm 1.79$ \\
I & 150/150 & $3.18 \pm 0.78$ & Z & 99/150 & $6.35 \pm 3.23$ \\
\bottomrule
\end{tabular*}
}
\par\smallskip
\parbox{\linewidth}{\footnotesize {Crossings: episodes reaching
the target's single-frame IoU threshold (\cref{sec:attainment-examples}),
out of 150 across three policies.
Time: mean capped first-crossing time $\pm$ sample std across three policy means
(50 episodes each), in seconds. Unsuccessful episodes contribute their
recorded end times. A dash indicates no crossing in any of the
150 episodes.}}
\end{table}

\begin{figure}[!t]
\color{black}
\centering
\includegraphics[width=\linewidth]{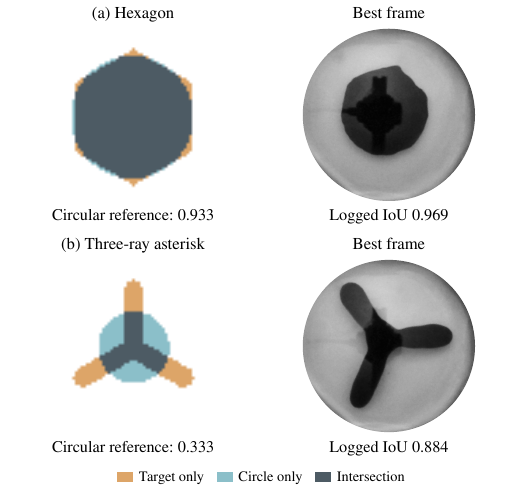}
\caption{{Area overlap and the interpretation of IoU. Left: a
filled circle positioned and sized to maximize overlap with the fixed target
mask, colored by intersection and unmatched regions. Right: selected ferrofluid
photographs with their logged single-frame IoU, computed using shape registration
in \cref{sec:learning-implementation}. A circle overlaps substantially with the
hexagon despite lacking corners, whereas its overlap with the three-ray
asterisk is lower.}}
\label{fig:circular-reference}
\end{figure}

\begin{figure*}[t]
\color{black}
\centering
\includegraphics[width=\textwidth]{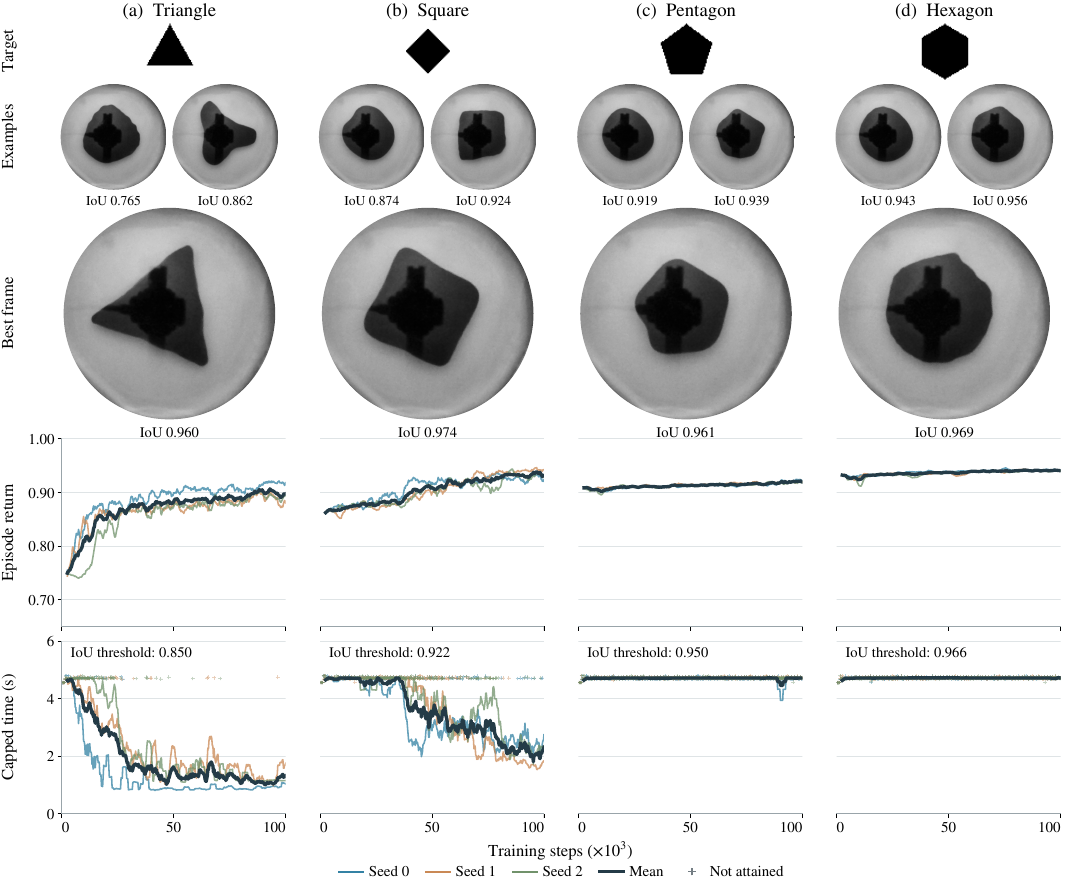}
\caption{{Polygon targets, example silhouettes, episode returns
and capped first-attainment times.
Each column shows the target, two smaller examples at lower and intermediate
IoUs, the best recorded frame, episode return, and capped first-attainment
time at the IoU threshold shown in each lower panel
(\cref{sec:physical-learning}).
Image labels report single-frame IoU. Thin curves show three seeds and the
bold curve their mean, with trailing 10-episode averages. Plus signs mark
unsuccessful episodes at their recorded end times. Each plot row shares a
vertical scale; horizontal axes count training steps in thousands, excluding
resets. Each policy trains for 100\,000 steps (400 episodes of 250 steps).}}
\label{fig:polygons-results}
\end{figure*}

\subsection{Polygons}
\label{sec:polygons}
The polygon experiments test whether the controller forms sides and corners
beyond a round outline. Each policy trains for $N=400$ episodes of $T=250$
steps, totaling 100\,000 training steps. \Cref{fig:polygons-results} shows
{the triangle, square, pentagon and hexagon targets, episode returns and
capped first-attainment times.}
The mean training {episode return} rises from 0.786 to 0.899 for the triangle and
from 0.868 to 0.933 for the square between the first and last 50 episodes.
The selected images show straight sides and distinct corners for both targets.

{The pentagon and hexagon show much smaller learning gains:
their mean training returns increase by only 0.014 and 0.012 between the first
and last 50 episodes. Final-policy {mean episode returns are} 0.916 and 0.941,
respectively, compared with 0.868 and 0.925 for the triangle and square.
The selected pentagon and hexagon outlines nevertheless have less distinct
sides and corners than their targets. Because IoU measures area overlap,
these geometric differences can coexist with high scores
(\cref{fig:circular-reference}).}

{The attainment curves separate these outcomes further.
Triangle crossings increase from 30/150 episodes in the first 50 episodes
per seed to 149/150 in the last 50, while mean capped time falls from
4.19 to 1.17~s. For the square, crossings increase from 1/150 to 137/150
and capped time {falls from approximately 4.7 to 2.2~s}. The pentagon and hexagon cross
their thresholds in only 2 and 1 of their 1\,200 training episodes,
respectively: their already high circular overlaps place $q$ close to one.}
{In final-policy evaluation, the triangle and square cross in
122/150 and 127/150 episodes, with mean capped times of 2.35 and 3.05~s;
the pentagon and hexagon have no crossings (\cref{tab:attainment-summary}).}

\begin{figure*}[t]
\color{black}
\centering
\includegraphics[width=\textwidth]{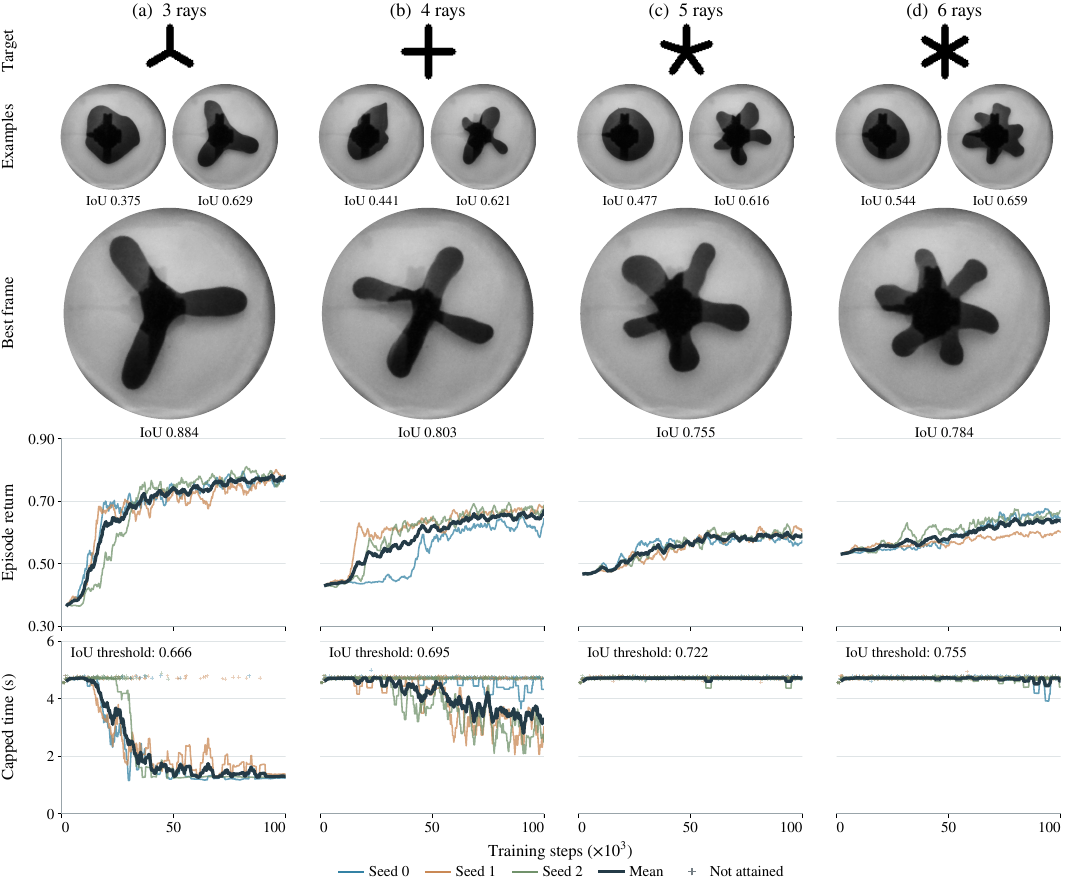}
\caption{{Three- through six-ray asterisks, using the image,
episode-return and capped first-attainment-time layout of \cref{fig:polygons-results}.
Each plot row shares a vertical scale. Curves use trailing 10-episode averages;
plus signs mark unsuccessful episodes. Each policy trains for 100\,000 steps
(400 episodes of 250 steps).}}
\label{fig:asterisk-results}
\end{figure*}

\subsection{Asterisks}
\label{sec:asterisks}
Asterisks test the formation of separated radial branches. Each policy
trains for $N=400$ episodes of $T=250$ steps, again totaling 100\,000
training steps. {The three-ray target shows the largest increase
in mean episode return, from 0.409 to 0.774 between early and late training;}
the four-, five- and
six-ray targets reach late-training means of 0.656, 0.590 and 0.638
(\cref{fig:asterisk-results}). The selected three- and four-ray images
show distinct branches, while the five- and six-ray examples have shorter
lobes around a larger central region.

Evaluation of the final policies gives mean {episode returns} of 0.658, 0.597,
0.529 and 0.613 for three through six rays. {Adding rays does
not produce a monotonic change in score.} The three-ray target also has the largest training--evaluation
difference: its final-policy mean is $0.658\pm0.073$, compared with a
late-training mean of $0.774\pm0.004$.
{Late-training returns reflect evolving stochastic policies,
whereas evaluation uses one frozen final checkpoint per run. Performance can
fluctuate during training, so a final checkpoint may perform below its run's
late-training average.}

{For the three-ray target, threshold crossings increase
from 2/150 early episodes to 149/150 late episodes, with capped time falling
from 4.69 to 1.32~s. Crossings within the first 2~s increase from 0/150 to
149/150, showing earlier shape formation by late training.
The four-ray target increases from 0/150 to 73/150,
with capped time {falling from approximately 4.7 to 3.4~s}. Five and six rays cross their
thresholds in only 2 and 17 of their 1\,200 training episodes. Visible lobes
can therefore remain below this IoU criterion when their lengths and central
area differ from the target.}
{Final policies cross in 98/150 and 28/150 episodes for three
and four rays, with mean capped times of 3.02 and 4.24~s.
Five rays have no crossings, while six rays cross in 5/150 episodes
(\cref{tab:attainment-summary}).}

\begin{figure*}[!p]
\color{black}
\centering
\includegraphics[width=0.945\textwidth]{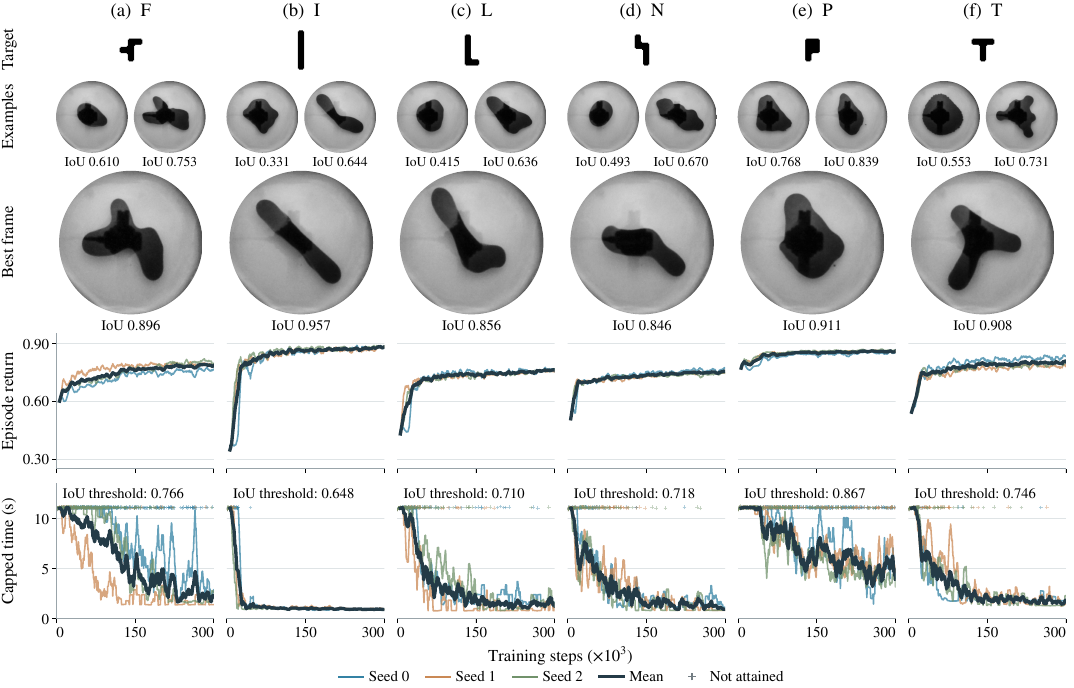}
\caption{{Pentomino targets F, I, L, N, P and T, using the
image, episode-return and capped first-attainment-time layout of
\cref{fig:polygons-results}. Curves use trailing 10-episode averages; plus signs mark
unsuccessful episodes. All 12 pentominoes share a vertical scale for each
plot row. Each policy trains for 300\,000 steps (500 episodes of 600 steps).}}
\label{fig:pentomino-results-one}
\par\vspace{0.5\baselineskip}
\includegraphics[width=0.945\textwidth]{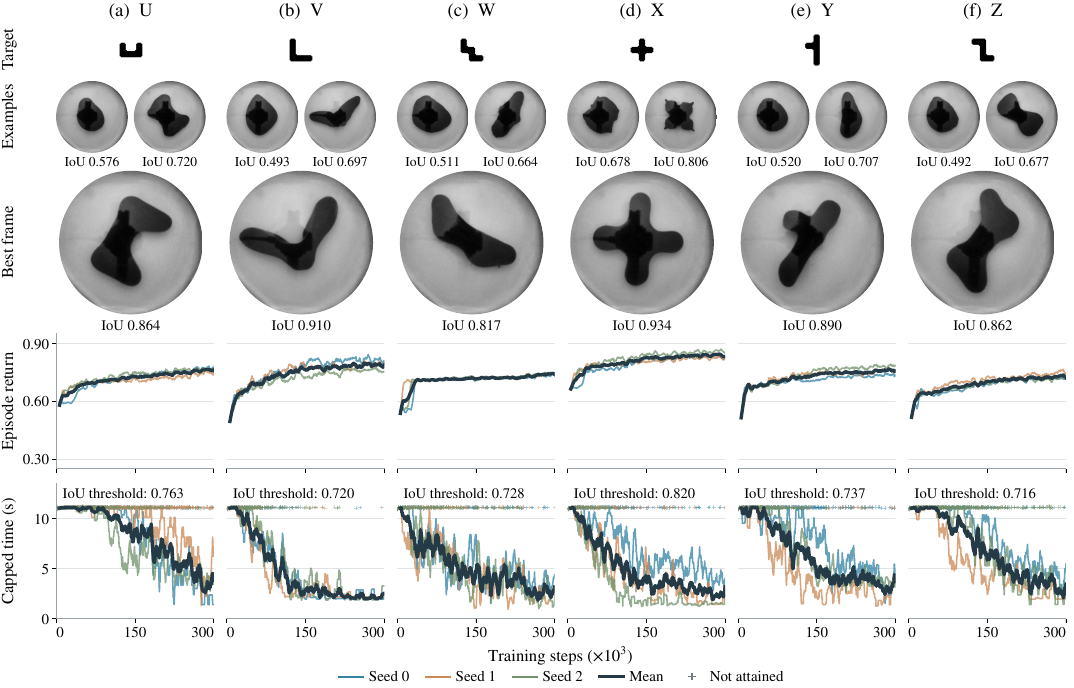}
\caption{{Pentomino targets U, V, W, X, Y and Z, continuing
\cref{fig:pentomino-results-one} with the same image, threshold rule and plotting
conventions.}}
\label{fig:pentomino-results-two}
\end{figure*}

\subsection{Pentominoes}
\label{sec:pentominoes}
Pentominoes extend the targets to bends, concavities and unequal branches.
Each policy trains for $N=500$ episodes of $T=600$ steps, totaling
300\,000 training steps. For the more geometrically complex pentomino
targets, we used longer episodes to allow more time for shape formation
and three times as many training steps to support policy learning. These
different training conditions should be considered when comparing
performance across target families.

{Mean episode return increases in all 36 runs between early and
late training, with late-training target means ranging from 0.729 for Z to 0.879}
for I (\cref{fig:pentomino-results-one,fig:pentomino-results-two}).
{Final-policy evaluation ranges from 0.701 for Z to 0.853 for I.}

The images reveal differences that the scores alone obscure. The selected
F, I, L, T, U, V, X, Y and Z frames reproduce the main limbs. In P, N and W, nearby limbs and corners merge: P appears
as a compact body with a protrusion, N as a kinked bar, and W as a bent
bar rather than a staircase. P nevertheless has a high evaluation score
of 0.849 because much of its target is compact. These examples show both
the variety of accessible outlines and the difficulty of resolving adjacent
features.

{All 12 pentominoes show more frequent crossings and
lower capped times between early and late training. Across targets, crossings
range from 0--77/150 in the first 50 episodes per seed and 116--150/150 in
the last 50. For example, the I pentomino's mean capped time falls from
6.35 to 0.95~s, and F's from 10.71 to 2.24~s. More frequent and earlier
crossings across full episodes support learned control of shape formation
beyond the selected photographs.}

{Some target shapes are attained more reliably than others.
With identical training step counts and episode lengths, the I and L
pentominoes cross their thresholds in 150/150 and 143/150 final-policy
evaluation episodes, respectively, compared with 52/150 for U
(\cref{tab:attainment-summary}). These differences suggest that producing a
range of shapes does not imply equally reliable control of every geometry.
The comparison also depends on the target-specific IoU criterion, which
captures some geometric features better than others.}

\subsection{\texorpdfstring{\textcolor{black}{Transient shape formation and persistence}}{Transient shape formation and persistence}}
\label{sec:pathways}
{The learned shapes appear during motion and then deform further.
\Cref{fig:episode-traces} follows the training episode containing the highest
recorded single-frame IoU for the three-ray asterisk. Current commands vary
across the full 0--0.5~A range as the silhouette develops. The IoU reaches
0.884 before falling below 0.49 within 0.15~s of its peak. The image sequence
shows that the high-scoring outline is a brief part of a longer deformation
trajectory.}

This distinction also appears in evaluation. For the I pentomino, the
mean single-frame peak is 0.878, whereas the mean IoU over all observations
is 0.407. We measure the camera-timestamp span of consecutive observations
containing each episode's first maximum and remaining within 0.05 IoU of it.
{\Cref{tab:near-peak-persistence} reports the average of the
three policies' median spans for each target: 0.13--0.34~s for polygons and
0.07--0.13~s for asterisks and pentominoes.}
These sampled spans describe persistence near each episode's own peak,
including imperfect matches, rather than uninterrupted time above a fixed
success threshold.

{The trajectories and persistence measurements show brief
near-peak configurations within ongoing deformation.
{Together with earlier threshold crossings during learning,
they characterize the timing and duration of transient shaping.}}

\FloatBarrier

\noindent\begin{minipage}[t]{\linewidth}
\color{black}
\centering
\includegraphics[width=\linewidth]{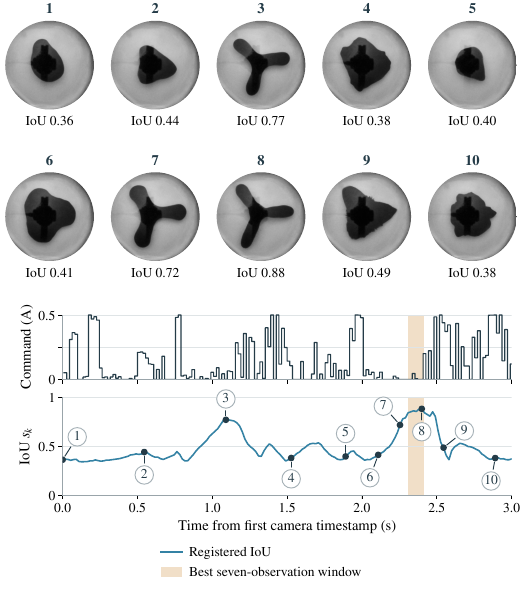}
\captionof{figure}{{Formation and decay of the three-ray
asterisk during training (seed 2, episode 214), shown through camera frames,
current commands and registered IoU. Numbered points on the IoU curve identify
the photographs, with peak IoU at frame 8 and shading marking the best
seven-observation window.}}
\label{fig:episode-traces}
\end{minipage}\par
\medskip
\noindent\begin{minipage}{\linewidth}
\captionof{table}{{Persistence within 0.05 IoU of each episode's peak}}
\label{tab:near-peak-persistence}
\centering
\footnotesize
\setlength{\tabcolsep}{2.5pt}
\renewcommand{\arraystretch}{0.98}
{\color{black}%
\begin{tabular*}{\linewidth}{@{\extracolsep{\fill}}lc|lc@{}}
\toprule
Target & Span (s) & Target & Span (s) \\
\midrule
Triangle & $0.13 \pm 0.01$ & L & $0.08 \pm 0.01$ \\
Square & $0.16 \pm (<0.01)$ & N & $0.11 \pm 0.02$ \\
Pentagon & $0.34 \pm 0.06$ & P & $0.13 \pm 0.01$ \\
Hexagon & $0.31 \pm 0.03$ & T & $0.12 \pm 0.03$ \\
3 rays & $0.09 \pm 0.02$ & U & $0.08 \pm 0.03$ \\
4 rays & $0.08 \pm 0.01$ & V & $0.08 \pm 0.01$ \\
5 rays & $0.13 \pm 0.03$ & W & $0.12 \pm 0.02$ \\
6 rays & $0.09 \pm 0.04$ & X & $0.08 \pm 0.01$ \\
F & $0.10 \pm 0.02$ & Y & $0.07 \pm 0.03$ \\
I & $0.09 \pm (<0.01)$ & Z & $0.08 \pm 0.03$ \\
\bottomrule
\end{tabular*}
}
\par\smallskip
\parbox{\linewidth}{\footnotesize {Mean $\pm$ sample {std} across
three policy medians, each over 50 final-policy evaluation episodes.}}
\end{minipage}\par
\endgroup

\Needspace{5\baselineskip}
\section{Conclusions}
\label{sec:conclusions}
{This work demonstrates how feedback policies learned
directly through physical interaction can exploit ferrofluid dynamics for
prescribed transient shaping with a single fixed electromagnet, offering a way
to explore control strategies for fluid-based soft robots.}
{We evaluate three
independently trained policies per target across 20 polygon, asterisk and
pentomino silhouettes. Improvement is consistent across training seeds:
each of the 60 trained policies achieves a higher mean evaluation return
than the untrained-policy average for its target.}
{Between early and late training, attainment frequency
increases and mean capped time decreases for the triangle, square, three-
and four-ray asterisks, and all 12 pentominoes.}
{These improvements across complete episodes show that
the policies learn to guide the moving ferrofluid toward prescribed transient
shapes.}

{IoU provides a simple way to define the shaping objective
from a drawn target silhouette, but area overlap has limitations in capturing
geometric detail. Attainment counts and times also depend on the chosen threshold.}
{The pentagon, hexagon and five-ray asterisk do not cross their thresholds in
final-policy evaluation. The image examples show recognizable five-ray
branching, whereas the pentagon and hexagon have poorly resolved corners.
Conversely, P, N and W pentominoes can cross their thresholds despite merged
limbs or rounded corners.
These differences motivate complementing IoU with measures and rewards that
account for contours, corners and branching structure.}

{More broadly, learning directly through physical
interaction offers an experimental tool for studying complex systems, opening
avenues for soft robotic designs that exploit their dynamics.} Varying interfacial tension and viscosity, and
exploring non-Newtonian ferrofluid formulations, would help determine how
material design affects the range, duration, and behavior of controllable deformations.
Additional electromagnets could expand that range through richer spatial
actuation and help resolve the closely spaced limbs of the P, N and W
pentominoes. Continual learning could support adaptation \citep{dohare2024},
while reduced-order models and optimal-transport methods could inform
predictive shape control
\citep{alora2025,solomon2015,lanzetti2026variational,chen2017transport,terpin2024dynamic}.
Prescribed transient shapes could also serve as visual elements in
reconfigurable ferrofluid displays. {Encapsulating the
ferrofluid in a soft membrane could enable soft robotic devices that use the
fluid's dynamics for controlled transient deformations to interact with their
surroundings.}

\appendices
\section{\texorpdfstring{\textcolor{black}{Additional related work}}{Additional related work}}
\label[appendix]{sec:related-work}
\subsection{\texorpdfstring{\textcolor{black}{Liquid robotics}}{Liquid robotics}}
{%
  In magnetic liquid robotics, the material and its interface determine which deformations a field can produce and whether they persist \citep{wang2024liquid,jahangir2026}. {Jamming of nonmagnetic nanoparticles at the interface can retain magnetically prepared ferrofluid shapes after field removal \citep{zhao2022}.} Individual ferrofluid droplets and their assemblies form mobile structures whose morphology depends on the imposed field \citep{fan2020deformabilities}; programmed rolling, wobbling and oscillating gaits provide different modes of locomotion \citep{yang2023multimodal}. Other magnetic suspensions extend pseudopodia for locomotion and engulfment \citep{zhang2023amoeba}.

Feedback control supports navigation and manipulation. Visual feedback enables planar steering \citep{probst2011}, and reduced motion models support predictive three-dimensional navigation \citep{skowronek2025}. {Position control accompanies field-commanded elongation for three-dimensional motion \citep{fan2023locomotion} and planar object manipulation \citep{kong2024}.} {A $12\times12$ coil array with visual position feedback supports droplet path tracking and coordinated manipulation of up to 72 ferrofluid droplets \citep{cui2025array}.} In levitation, feedback instead regulates the optically measured gap between a droplet apex and an overhead electromagnet \citep{ohji2020}. {Joint control of electrode voltages and electromagnet currents drives an electrolyte through Lorentz forces. A model-predictive controller uses a simulation-trained Koopman predictor to steer the flow toward prescribed velocity and vorticity fields \citep{uchytil2025}.}

{Liquid robotic systems also use the fluid to manipulate payloads or probe surrounding materials.} Visual feedback positions nonmagnetic solid particles by deforming their supporting air--ferrofluid interface \citep{cenev2021}. {Coil arrays guide mobile permanent magnets beneath ferrofluid droplets to coordinate droplet sorting and an automated biochemical assay \citep{yu2020ferrobotic}.} Measured droplet deformation estimates the viscosity or elasticity of surrounding media \citep{wang2026tweezer}. These platforms give magnetic liquids distinct roles as mobile bodies, actuation media and mechanical probes, with control and measurement tailored to each role.
}

\par\medskip
\noindent\begin{minipage}[t]{\linewidth}
    {\color{black}
      \section{Attainment threshold examples}
      \label[appendix]{sec:attainment-examples}}
    \centering
\includegraphics[width=\linewidth]{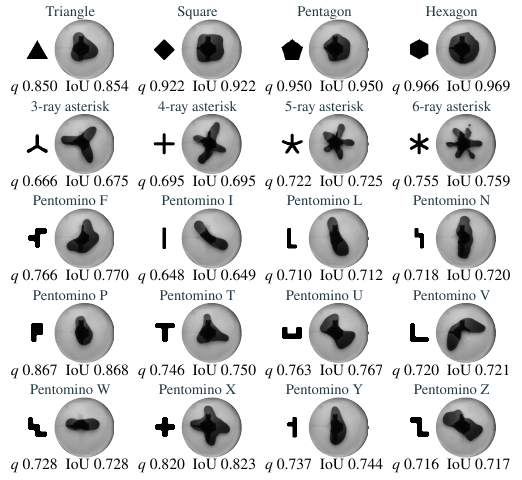}
\captionof{figure}{{Target masks and saved training photographs just above
the unrounded thresholds defined in \cref{sec:physical-learning}. Labels give $q$
and each frame's IoU, rounded down to three decimals.}}
\label{fig:attainment-examples}

\end{minipage}\par

\par\medskip
\noindent\begin{minipage}[t]{\linewidth}
  \section{\texorpdfstring{\textcolor{black}{Threshold sensitivity}}{Threshold sensitivity}}
  \label[appendix]{sec:threshold-sensitivity}
  {We vary $\alpha$ in $q=D+\alpha(1-D)$ using the same
evaluation episodes and single-frame criterion. The final-policy crossing
rate changes substantially, while remaining above the untrained-policy rate
at all three settings (\cref{tab:threshold-sensitivity}).}

\par\medskip
\centering
\captionof{table}{{Evaluation crossings at three threshold settings}}
\label{tab:threshold-sensitivity}
\footnotesize
\setlength{\tabcolsep}{3pt}
\renewcommand{\arraystretch}{1.05}
\begin{tabular*}{\linewidth}{@{\extracolsep{\fill}}ccc@{}}
\toprule
{$\alpha$} & {Untrained policy} & {Final policy} \\
\midrule
0.25 & 28/3\,000 (0.9\%) & 2\,611/3\,000 (87.0\%) \\
0.50 & 0/3\,000 (0.0\%) & 1\,759/3\,000 (58.6\%) \\
0.75 & 0/3\,000 (0.0\%) & 184/3\,000 (6.1\%) \\
\bottomrule
\end{tabular*}

\par\smallskip
\parbox{\linewidth}{\footnotesize{Counts pool 150 episodes
per target across 20 targets. The main analysis uses $\alpha=0.5$.}}

\end{minipage}\par

\vspace{0.75\baselineskip} %
\begingroup
\let\manuscriptbibliography\thebibliography
\renewcommand{\thebibliography}[1]{\manuscriptbibliography{#1}\interlinepenalty=10000}
\IEEEtriggeratref{41} %

\endgroup

\end{document}